\documentclass[letterpaper, 10 pt, conference]{ieeeconf} 
\IEEEoverridecommandlockouts  
\usepackage{cite}
\usepackage{amsmath,amssymb,amsfonts}
\usepackage{graphicx}
\usepackage{hyperref}
\usepackage{tabularx} 
\usepackage{rotating} 
\usepackage{booktabs}

\hypersetup{
    colorlinks=true,
    linkcolor=magenta,
    filecolor=magenta,      
    urlcolor=magenta,
}

\title{\LARGE \bf Thermal-NeRF: Neural Radiance Fields from an Infrared Camera}  

\author{Tianxiang Ye$^{1}$, Qi Wu$^{1}$, Junyuan Deng$^{1}$, Guoqing Liu$^{1}$, \\ Liu Liu$^{2}$, Songpengcheng Xia$^{1}$, Liang Pang$^{3}$, Wenxian Yu$^{1}$ and Ling Pei$^{1*}$
\thanks{This work was supported in part by the National Nature Science Foundation of China (NSFC) under Grant No.62273229, No.61873163 separately and in part by smart City beidou spatial-temporal digial base construction and application industrialization(HCXBCY-2023-020).}
\thanks{$^{1}$Shanghai Key Laboratory of Navigation and Location Based Services, Shanghai Jiao Tong University, $^{2}$Hefei University of Technology, $^{3}$Shanghai Slamtec Research.}%
\thanks{*Corresponding author: Pei Ling {\tt\small ling.pei@sjtu.edu.cn}}%
}

\begin{document}

\maketitle
\thispagestyle{empty}
\pagestyle{empty}

\begin{abstract}
In recent years, Neural Radiance Fields (NeRFs) have demonstrated significant potential in encoding highly-detailed 3D geometry and environmental appearance, positioning themselves as a promising alternative to traditional explicit representation for 3D scene reconstruction. However, the predominant reliance on RGB imaging presupposes ideal lighting conditions—a premise frequently unmet in robotic applications plagued by poor lighting or visual obstructions. This limitation overlooks the capabilities of infrared (IR) cameras, which excel in low-light detection and present a robust alternative under such adverse scenarios. To tackle these issues, we introduce Thermal-NeRF, the first method that estimates a volumetric scene representation in the form of a NeRF solely from IR imaging. By leveraging a thermal mapping and structural thermal constraint derived from the thermal characteristics of IR imaging, our method showcasing unparalleled proficiency in recovering NeRFs in visually degraded scenes where RGB-based methods fall short. We conduct extensive experiments to demonstrate that Thermal-NeRF can achieve superior quality compared to existing methods. Furthermore, we contribute a dataset for IR-based NeRF applications, paving the way for future research in IR NeRF reconstruction, see \url{https://github.com/Cerf-Volant425/Thermal-NeRF}.
\end{abstract}

\section{INTRODUCTION}
Over the past decades, image-based 3D reconstruction has not only achieved remarkable success but also emerged as a pivotal area of research within the field of computer vision\cite{ma2018review}. Its high accuracy, efficiency, and scalability have led to widespread applications across numerous domains, including artificial intelligence, robotics, autonomous driving, and virtual reality. Despite these advances, reconstructing 3D scenes within indoor environments presents unique challenges, due to the complexity of accurately modeling the geometry, spatial positioning, topological relationships, and semantic properties of these settings\cite{kang2020review}. This challenge is critical, as overcoming it could significantly enhance tasks reliant on precise indoor positioning, such as semantic segmentation, scene understanding, and environmental perception. 

The pressing need to address these intricate challenges has opened avenues for the exploration of innovative approaches. Neural Radiance Fields (NeRFs) have recently risen to prominence for tasks like 3D scene representation and novel view synthesis\cite{mildenhall2021nerf}. NeRF combine a multilayer perceptron (MLP) for learning spatial information with differential rendering to realistically simulate light interactions within a scene.  It marks a departure from explicit representation methods like voxel, point cloud, and mesh, adopting an implicit approach that excels in capturing detailed scene nuances\cite{engel2017direct, curless1996volumetric, hoppe1993mesh}. Up to this date, NeRF has demonstrated effective performance under ideal conditions, characterized by stable and sufficient lighting and free of visual occlusions, with its efficacy proven on high-quality, real-world images captured in such optimal environments. However, real-world environments frequently deviate from the premise of photometric consistency, exhibiting challenges such as variable illumination and low light conditions. These factors significantly impair the performance of RGB-based NeRF approaches. Despite efforts to optimize NeRF for these conditions\cite{mildenhall2022nerf, martin2021nerf, fujitomi2022lb}, such adaptations have fallen short in the face of severe visual degradation.

In the particularly demanding contexts of fire incidents and nighttime operations, the RGB cameras that feed NeRF with data fail to function effectively, rendering the system unsuitable. This highlights the necessity for infrared (IR) imaging approach\cite{vollmer2021infrared} that can operate reliably in such challenging environments. IR cameras, with their ability to capture thermal signatures rather than relying on visible light, offer a distinct advantage in scenarios where conventional cameras falter. They excel in penetrating through smoke, detecting heat sources in the dark, and providing clear images despite the presence of obstructions, making them indispensable for scene reconstruction where darkness prevails and thermal sensitivity is crucial. However, the inherent characteristics of IR images, such as low contrast, sparse features and limited textures, which result in subtle pixel-level variations\cite{ko2023large, kuang2019single, liu2015general}. Since NeRF primarily constrains pixel-level loss, the nuanced variations in IR images significantly hinder its ability to accurately reconstruct scenes, posing challenges to NeRF reconstruction.

Given these learnings, we propose Thermal-NeRF, an innovative approach that tackles NeRF estimation from IR cameras, enabling the scene reconstruction from visually degraded environments. We apply a thermal mapping to model IR images' thermal value, ensuring the consistency in heat representation. And we introduce a structural thermal constraint to harness the structural information within images, offering vital constraints for IR images marked by sparse features and textures. Our dense experiments demonstrate that our method outperforms existing approaches in reconstruction quality. Furthermore, considering the current scarcity of IR datasets, we have compiled a targeted dataset for IR-based NeRF reconstruction. To summarise, the primary technical contributions are as follows:
\begin{enumerate}
    \item We propose Thermal-NeRF, to the best of our knowledge, our approach is the first attempt to tackle NeRF estimation from IR cameras in situations where RGB cameras are inadequate.  
    \item We propose thermal mapping for modeling IR thermal values and and a structural thermal constraint on thermal distribution, which leverages IR's distinctive features for high-fidelity 3D representations.
    \item We build an IR dataset featuring visually degraded scenes to validate our proposal and address the gap in existing IR datasets for NeRF. We will release the dataset to establish a benchmark for future work.  
\end{enumerate}

\section{RELATED WORK}
In this section, we examine key related works and explore their connections to our proposed method.
\subsection{3D Scene Representation}
Addressing 3D scene representation has been a longstanding challenge in the fields of computer vision and computer graphics. Existing methods such as depth maps\cite{newcombe2011dtam}, point clouds\cite{engel2017direct}, voxel grids\cite{curless1996volumetric}, and meshes\cite{hoppe1993mesh} have made significant contributions to this domain. However, each of these approaches comes with inherent limitations, be it in terms of resolution, computational efficiency, or the ability to capture complex geometries.

In contrast, the recent introduction of implicit coordinate-based representations\cite{lombardi2019neural,park2019deepsdf,mildenhall2021nerf, deng2023nerf}, exemplified by NeRF\cite{mildenhall2021nerf}, represents a significant advancement. NeRF models a scene through a MLP as a continuous function of scene radiance and volume density, learned from a set of 2D RGB images. At test time, NeRF can render novel views from arbitrary 3D camera positions and viewing angles. However, a common limitation in most NeRF-based research is the need for high-quality input scene data, assuming ideal conditions for optimal performance. 

On one hand, efforts have been made to enhance NeRF's adaptability to different lighting scenarios, RawNeRF\cite{mildenhall2022nerf} functions effectively in low-light environments, W-NeRF\cite{martin2021nerf} is designed to cope with environments that exhibit changing lighting conditions and mitigates the effects of dynamic objects within a scene, LB-NeRF\cite{fujitomi2022lb} manages scenes with transparent medium, overcoming challenges posed by light refraction, making NeRF more robust to light and broadening NeRF's applicability. On the other hand, \cite{wang2021neus,kulkarni2022directed,metzer2023latent,xiao2022resnerf,tang2023delicate} have seen significant improvements, as evidenced by various studies. Neus\cite{wang2021neus}, for example, achieves high-fidelity reconstruction of objects and scenes from 2D RGB images, while ResNeRF\cite{xiao2022resnerf} excels in synthesizing novel views with high fidelity, maintaining 3D structures in large-scale indoor scenes. Other researches\cite{metzer2023latent,tang2023delicate} focus on enhancing object textures, resulting in improved surface mesh quality.

Despite these developments, all these methods predominantly rely on traditional RGB camera inputs, which are less effective in dark or smoke-filled environments. Such challenging conditions are not only common but also crucial in the context of indoor scene reconstruction, underscoring a critical gap in the current state of the art.

\subsection{Infrared Cameras}
Infrared cameras\cite{vollmer2021infrared,bao2023heat}, harnessing radiation in the infrared spectrum, particularly between 3 to 14 $\mu m$  wavelengths, produce spatial temperature distribution maps, independent of external light sources, capturing the infrared radiation emitted by objects. This technology has seen wide civilian application\cite{he2021infrared}, ranging from fever scanners to insulation, due to reduced costs and enhanced portability. In addition, their high sensitivity has opened doors to various optical applications like fire prediction, electrical hotspot detection, and nighttime monitoring\cite{he2020infrared}, offering advantages over RGB cameras, especially in distinguishing between objects based on heat signatures. 

However, in the realm of IR scene reconstruction, distinct challenges are encountered, primarily attributed to the inherent properties of IR images. These images are typically characterized by sparsity in detail and diminished contrast\cite{vollmer2021infrared}. Moreover, the acquisition of camera parameters, with a specific emphasis on extrinsic variables, proves infeasible through standard Structure from Motion (SfM) methodologies that are predicated on feature matching.  Notably, techniques like COLMAP\cite{schoenberger2016mvs}, which have found widespread application in NeRF for pose estimation, encounter limitations in this context. To tackle these issues, contemporary research has involved the fusion of IR with other modalities\cite{hou2020vif, ma2019visible, lang20083d, poggi2022cross}, compensating for the spatial data limitations of IR images. For example, Ma et al.\cite{ma2019visible} integrates RGB images to enrich the feature set and enhance depth perception, and Lang et al. \cite{lang20083d} merges with Inertial Measurement Unit (IMU) data to provide pose information. Also, several studies have capitalized on the unique aspects of IR imaging using deep learning methods\cite{kuang2019single, poggi2022cross,ko2023large}, concentrating on augmenting IR-specific features. Notably, X-NeRF\cite{poggi2022cross} is the sole existing NeRF-based approach that concerns IR images, it creates a cross-spectral scene representation and learns the relative poses between IR and RGB sensors. In contrast, our approach is distinguished by its exclusive reliance on the IR modality for NeRF without auxiliary modalities.

\begin{figure*}[thpb]
    \centering
    \includegraphics[width=0.95\linewidth]{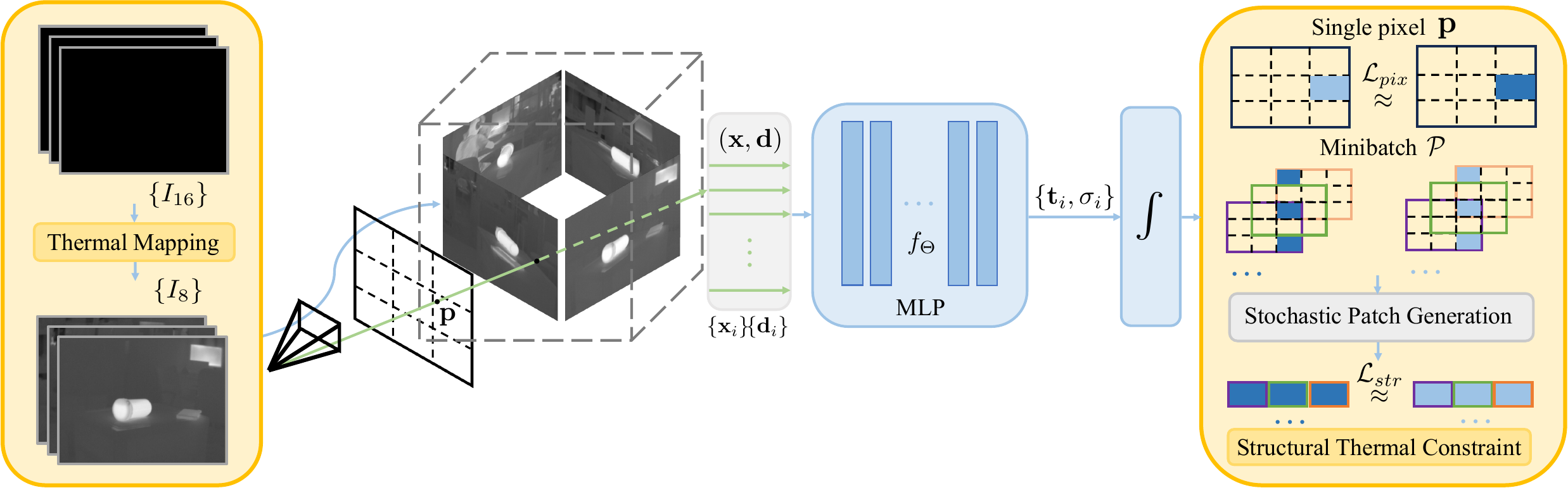}
    \caption{Overview of the proposed Thermal-NeRF. Initially, a set of 16-bit IR images $\{I_{16}\}$, undergoes thermal mapping to be transformed into 8-bit images $\{I_8\}$. The method includes a scene contraction step, which compresses the indoor space into a predefined, fixed-size bounding box. Utilizing the camera parameters, ray bundles are generated through the contracted indoor scene. These bundles are then sampled to yield sampling points and directions $\{\mathbf{x}_i, \mathbf{d}_i\}$. The samples are encoded and fed into the MLP $f_{\Theta}$. This step aggregates the output radiance $\mathbf{t}_i$ and densities $\mathbf{\sigma}_i$ to compute the thermal value. A unique structural thermal constraint is proposed to optimize the loss within mini-patches formed by stochastic pixels, see Equation \ref{eq10}.}
    \label{model}
\end{figure*}

\section{METHODOLOGY}

In this section, we introduce Thermal-NeRF, a new framework for scene representation with IR images. We first briefly introduce the background of original NeRF (Sec. \ref{bg}). Then we implement thermal mapping (Sec. \ref{tm}) to model thermal values and finally we innovate a structural thermal constraint (Sec. \ref{tl}) to leverage structural information. The overall framework is illustrated in Fig. \ref{model} and described in detail below. 

\subsection{Background: Neural Radiance Field}
\label{bg}
NeRF\cite{mildenhall2021nerf} optimizes a neural radiance field parameterized by an MLP network $f_{\Theta}:(\mathbf{x},\mathbf{d})\rightarrow(\mathbf{c},\mathbf{\sigma})$, which maps Cartesian input coordinates $\mathbf{x}\in\mathbb{R}^3$ and viewing direction $\mathbf{d}\in\mathbb{S}^2$ to the predicted color $\mathbf{c}\in\mathbb{R}$ and volume density $\sigma\in\mathbb{R}$. 

Considering the errors introduced by applying the sensor calibration and readout delay of the data transmission, we apply pose refinement to optimize the camera pose and the neural field simultaneously to improve
the image quality\cite{katragadda2023nerf, wang2021nerf}. We optimize camera views jointly with scene representation through an $SE(3)$ transformation, the camera views are then utilized to create ray bundles. 
To render an image from the NeRF model, the color at each pixel $\mathbf{p}\in\mathbb{Z}^2$ on the image is obtained by volume rendering, aggregating the radiance along a ray $\mathbf{r}$ shooting from the camera position $\mathbf{o}_i$, passing through the pixel $\mathbf{p}$ into the volume\cite{eskicioglu1995image}
\begin{equation}
\hat{\boldsymbol{C}}(\mathbf{p})=\int_{h_n}^{h_f}T(h)\sigma(\mathbf{r}(h))\boldsymbol{c}(\mathbf{r}(h), \mathbf{d})dh,
\end{equation}
where $T(h) = \exp(-\int_{h_n}^h\sigma(s)ds)$ denotes the accumulated transmittance along the ray, and $\mathbf{r}(h) = \mathbf{o} + h\mathbf{d}$ denotes the camera ray that starts from camera origin $\mathbf{o}$ and passes through $\mathbf{p}$, with near and far bounds $h_n$ and $h_f$. 

The original NeRF minimizes a least squares error between the rendered prediction colors $\hat{C}(\mathbf{p})$ and ground truth colors $C(\mathbf{p})$ provided by the images. Akin to NeRF, Thermal-NeRF minimizes the error between rendered prediction thermal values $\hat{T}(\mathbf{p})$ and ground truth thermal values $T(\mathbf{p})$, which is detailed in the next section. 

\subsection{Thermal Mapping}
\label{tm}
Diverging from RGB imaging, IR imaging encapsulates thermal values at each pixel, reflecting temperature variations vital for accurately interpreting thermal scenes. The data for our Thermal-NeRF are a set of IR images $\{I_k\}_{k=1}^{N}$ in 16-bit single-channel format, where each pixel corresponds to a thermal value. The transformation of the thermal value of $T_{16}$ at a given 16-bit  pixel $\mathbf{p}$ for camera ray $\mathbf{r}$ can be formulated as a linear mapping\cite{vollmer2021infrared}, the final conversion is given by the equation
\begin{equation}T_{16}(\mathbf{p}) = \frac{\mathbf{p}}{\mathbf{k}} + \mathbf{b}, 
\label{eq1}\end{equation}
where $\mathbf{b}$ and $\mathbf{k}$ are two fixed imaging coefficients indicated by the IR camera.

Then we apply min-max scaling to the extreme thermal values within each data sequence $\{I_k\}_{k=1}^{N}$, ensuring heat consistency across all IR images, this approach also maximizes the contrast of images. To further aid visual task performance, we convert thermal values to 8-bit format, the conversion is written as 
\begin{equation}
T(\mathbf{p}) = \frac{T_{16}(\mathbf{p}) - \mathop{\min}_{x \in \mathcal{P}} T_{16}(x)} {\mathop{\max}_{x \in \mathcal{P}} T_{16}(x) - \mathop{\min}_{{x \in \mathcal{P}}} T_{16}(x)}255,
\label{eq2}\end{equation}
where $\mathcal{P}$ represents the set comprising every pixel within the data sequence, with $255$ signifying the range from $0$ to $255$ in an 8-bit format.

The rendered thermal value $\hat{T}(\mathbf{p})$ can then be compared against the corresponding ground truth thermal value $T(\mathbf{p})$, for all the pixels $\mathcal{P}$. We perform a least squares minimization of the pixel-wise thermal loss
\begin{equation}
\mathcal{L}_{pix}{(\Theta)=\frac1{\|\mathcal{P}\|}\sum_{\mathbf{p}\in\mathcal{P}}\|\hat{T}(\mathbf{p})-T(\mathbf{p})\|^2}.
\label{eq3}\end{equation}
While pixel-wise thermal loss plays a crucial role in refining details at the pixel level, its focus on high-frequency changes limits its effectiveness in capturing the subtler, low-contrast nuances of IR imaging\cite{hore2010image}, exploring structural constraints becomes essential to accurately capture the spatial and textural nuances of thermal scenes.

\subsection{Structural Thermal Constraint}
\label{tl}
Addressing the challenge of sparse textures and feature scarcity in IR imaging, where thermal radiation distribution is uneven, it's essential to focus on structural information. Before introducing the structural thermal constraint, we first need to define the evaluation metric based on thermal values to accurately assess structural information. To this end, we leverage Structural Similarity (SSIM)\cite{wang2004image} index, which adeptly captures the structural similarities or differences in images, and allows for localized assessment focusing on areas with concentrated thermal radiation. SSIM combines the three components of luminance, structure and contrast to comprehensively measure image quality. Let $X$ be the reference image and $Y$ be the test image, which is described as follows
\begin{equation}
    \mathrm{SSIM}(X,Y)=\frac{(2\mu_X\mu_Y+C_1)(2\sigma_{XY}+C_2)}{({\mu_X}^2+{\mu_Y}^2+C_1)({\sigma_X}^2+{\sigma_Y}^2+C_2)},
\end{equation}
where $\mu$ and $\sigma$ denote the mean and standard deviation respectively, and $\sigma_{XY}$ is the cross-correlation between $X$ and $Y$. $C_1$ and $C_{2}$ two constants, are stable coefficients when the mean value and the variance are close to zero. 

Additionally, IR imaging primarily detects variations in thermal values, reflecting temperature disparities rather than the brightness levels typically associated with RGB image quality assessment, so we should remove the luminance part. Then the heat-based metric HSSIM can be written as
\begin{equation}
\mathrm{HSSIM} = (X,Y)=\frac{2\sigma_{XY}+C_2}{\sigma_{X}^2+\sigma_{Y}^2+C_2},
\label{eq8}\end{equation}
we set $C_2 = 9 \times 10^{-4}$ in our work. Accounting for global information variability, we compute the mean HSSIM using local statistics derived from a sliding window $W$ with kernel size $l \times l$, moving across the image with stride $s$. Local statistics are assessed within each window to compute HSSIM, with the final metric being the average of these values.

During stochastic training of NeRF, the randomly sampled pixels in a minibatch $\mathcal{P}$ do not constitute a coherent local patch, leading to a total loss of their spatial interrelation. As suggested by \cite{xie2023s3im}, for each sampled minibatch of pixels $\mathcal{P}$, they can form a stochastic patch through a patch generation function $G(\mathcal{P})$, which is initialized randomly each time when called. In this way, the structural loss can capture the non-local structural thermal information across all the training images. Let $\mathcal{T}$ and $\hat{\mathcal{T}}$ be the sampled pixels transformed to the format of thermal values, where $\mathcal{T} = \{T(\mathbf{p})| \mathbf{p} \in \mathcal{P}\}$ and $\hat{\mathcal{T}} = \{\hat{T}(\mathbf{p})| \mathbf{p} \in \mathcal{P}\}$. The corresponding patches are generated as $G(\mathcal{T})$ and $G(\hat{\mathcal{T}})$ respectively. Typically, areas with higher temperatures tend to exhibit more detailed features, implying that pixel intensity can reflect thermal levels. This suggests that the intensity of a pixel can serve as a gauge for the thermal target. Hence, we define $E(\mathcal{P})$, which represents the expected thermal intensity within a patch, to compute the average thermal intensity
\begin{equation}
E(\mathcal{P})=\frac{1}{h\times w}\sum_{i=1}^{h\times w}T(\mathbf{p}_i),
\label{eq9}\end{equation}
where h, w denotes the height and width of the formed patch. The intensity within this range is confined between 0 and 1, which is conveniently used as a weighting factor for the structural loss function. Then the structural loss can be written as
\begin{equation}
\mathcal{L}_{str}(\Theta) = E(\mathcal{P})(1 - \mathrm{HSSIM}(G(\mathcal{T}), G(\hat{\mathcal{T}}))).
\label{eq10}\end{equation}
At this point, we can obtain the total loss, which is a combination of 
\begin{equation}
\mathcal{L}_{tot} = \mathcal{L}_{pix} + \mathcal{L}_{str}.
\label{eq11}\end{equation}

\subsection{Implementation Details}
Our code is based on nerfacto model proposed by nerfstudio\cite{nerfstudio} applying the thermal mapping and thermal constraint based on heat information. We train the model for $3 \cdot 10^4$ iterations on one NVIDIA RTX 3090 with the default optimizer and hyper-parameters as in nerfstudio, the training usually converges in about 15 to 20 minutes. And we set the sliding window's kernel size to $4$ and the stride to $4$, enabling us to compute the mean HSSIM without overlapping pixels.

\section{EXPERIMENTAL RESULTS AND ANALYSIS}
In this section, we validate the efficacy of Thermal-NeRF through experiments on a self-collected IR dataset of indoor environments, focusing on novel view synthesis and 3D object reconstruction. We conduct a detailed comparison of Thermal-NeRF with representative methods including the original NeRF\cite{mildenhall2021nerf}, Mip-NeRF 360\cite{barron2022mip}, and DVGO\cite{sun2022improved}.

\begin{figure}[b]
    \centering
    \includegraphics[width=0.99\columnwidth]{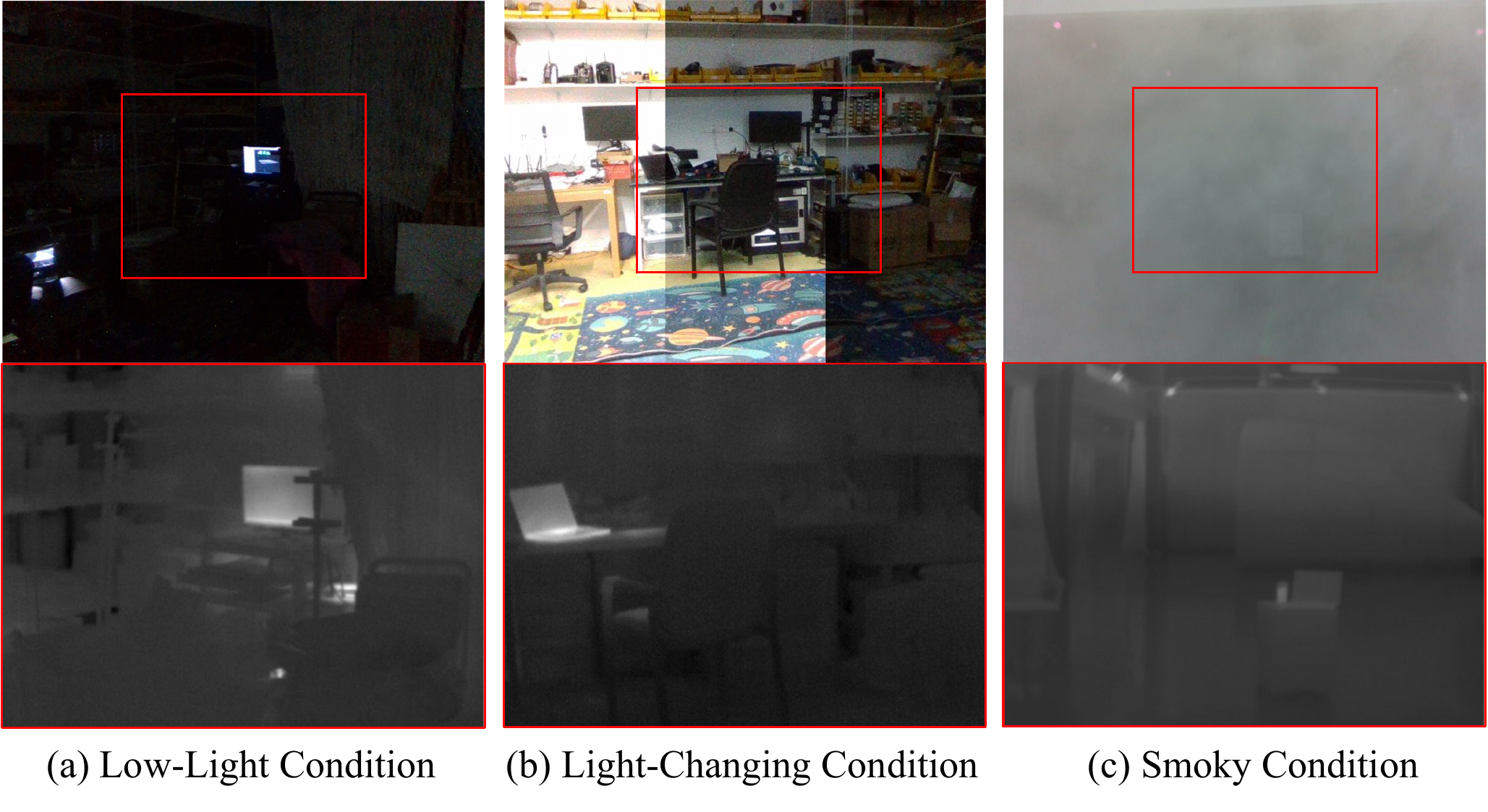}
    \vspace{-5mm}
    \caption{This illustration highlights that our actual sequences were documented in demanding conditions, including low and fluctuating lighting, as well as smoke. The RGB images, taken by camera RealSense D435, demonstrate the visual outcomes. The frames marked in red represents the same area as captured by both IR and RGB cameras, albeit with differing resolutions and fields of view.}
    \label{data_ex}
\end{figure}

\begin{table*}[t]
    \caption{Quantitative Evaluation on Self-collected Sequences Under Low-Light Conditions of Different Models}
    \label{table2}
    \centering
    \setlength{\tabcolsep}{3pt} 
    \begin{tabular}
    {@{} >{\raggedright\arraybackslash}m{3.5cm} m{1.4cm}<{\centering} m{1.4cm}<{\centering} m{1.4cm}<{\centering} m{1.4cm}<{\centering} m{1.4cm}<{\centering} m{1.4cm}<{\centering} m{1.4cm}<{\centering} m{1.4cm}<{\centering} @{} m{1.4cm}<{\centering} @{}}
         \toprule
        & \multicolumn{3}{c}{Sequnce 1} & \multicolumn{3}{c}{Sequence 2} & \multicolumn{3}{c}{Sequence 3} \\ 
        & \multicolumn{3}{c}{train/test views: 243/26} & \multicolumn{3}{c}{train/test views: 186/20} & \multicolumn{3}{c}{train/test views: 187/20}\\
        \cmidrule(lr){2-4} \cmidrule(lr){5-7} \cmidrule(l){8-10}
        
        Model & PSNR $\uparrow$ & SSIM $\uparrow$ & LPIPS $\downarrow$ & PSNR $\uparrow$ & SSIM $\uparrow$ & LPIPS $\downarrow$ &  PSNR $\uparrow$ & SSIM $\uparrow$ & LPIPS $\downarrow$ \\ 
        \midrule
        NeRF\cite{mildenhall2021nerf} & 27.57 & 0.83 & 0.47 & 21.32 & 0.82 & 0.45 & 26.22 & 0.75 & 0.26 \\
        Mip-NeRF 360\cite{barron2022mip} & 27.96 & 0.84 & 0.45 & 20.92  & 0.82 & 0.44 & 26.15 & 0.75 & 0.24 \\
        DVGO\cite{sun2022improved} & 19.18 & 0.79 & 0.57 & 13.16 & 0.71 & 0.63 & 15.83 & 0.65 & 0.41 \\
        Thermal-NeRF w/o pose\cite{wang2021nerf} & 28.19 & 0.88 & 0.37 & 17.60 & 0.85 & 0.37 & 25.93 & 0.77 & 0.21 \\
        Thermal-NeRF & \bf{32.41} & \bf{0.90} & \bf{0.36} & \bf{29.79} & \bf{0.88}  & \bf{0.37} & \bf{30.41} & \bf{0.79} & \bf{0.20} \\
        \bottomrule
    \end{tabular}
    \label{tab1}
\end{table*}

\begin{table*}[t]
    \caption{Quantitative Evaluation on Self-collected Sequences Under Light-Changing Conditions of Different Models}
    \label{table3}
    \centering
    \setlength{\tabcolsep}{3pt} 
    \begin{tabular}
    {@{} >{\raggedright\arraybackslash}m{3.5cm} m{1.4cm}<{\centering} m{1.4cm}<{\centering} m{1.4cm}<{\centering} m{1.4cm}<{\centering} m{1.4cm}<{\centering} m{1.4cm}<{\centering} m{1.4cm}<{\centering} m{1.4cm}<{\centering} @{} m{1.4cm}<{\centering} @{}}
         \toprule
        & \multicolumn{3}{c}{Sequnce 1} & \multicolumn{3}{c}{Sequence 2} & \multicolumn{3}{c}{Sequence 3} \\ 
        & \multicolumn{3}{c}{train/test views: 195/21} & \multicolumn{3}{c}{train/test views 213/23} & \multicolumn{3}{c}{train/test views 158/17} \\
        \cmidrule(lr){2-4} \cmidrule(lr){5-7} \cmidrule(l){8-10}
        
        Model & PSNR $\uparrow$ & SSIM $\uparrow$ & LPIPS $\downarrow$ & PSNR $\uparrow$ & SSIM $\uparrow$ & LPIPS $\downarrow$ &  PSNR $\uparrow$ & SSIM $\uparrow$ & LPIPS $\downarrow$ \\ 
        \midrule
        NeRF\cite{mildenhall2021nerf} & 27.62 & 0.83 & 0.48 & 27.48 & 0.76 & 0.25 & 30.57 & 0.84 & 0.39 \\
        Mip-NeRF 360\cite{barron2022mip} & 27.63 & 0.82 & 0.46 & 27.54 & 0.76 & 0.24 & 31.08 & 0.84 & 0.38 \\
        DVGO\cite{sun2022improved} & 18.69 & 0.78 & 0.63 & 18.50 & 0.68 & 0.36 & 20.86 & 0.74 & 0.50 \\
        Thermal-NeRF w/o pose\cite{wang2021nerf} & 28.30 & 0.89 & 0.40 & 27.09 & 0.87 & 0.22 & 29.69 & 0.91 & 0.38 \\
        Thermal-NeRF & \bf{31.27} & \bf{0.89} & \bf{0.38} & \bf{32.89} & \bf{0.89} & \bf{0.22} & \bf{34.17} & \bf{0.93} & \bf{0.37}\\
        \bottomrule
    \end{tabular}
    \label{tab2}
\end{table*}

\begin{table*}[ht]
    \caption{Quantitative Evaluation on Self-collected Sequences Under Smoky Conditions of Different Models}
    \label{table1}
    \centering
    \setlength{\tabcolsep}{3pt} 
    \begin{tabular}
    {@{} >{\raggedright\arraybackslash}m{3.5cm} m{1.4cm}<{\centering} m{1.4cm}<{\centering} m{1.4cm}<{\centering} m{1.4cm}<{\centering} m{1.4cm}<{\centering} m{1.4cm}<{\centering} m{1.4cm}<{\centering} m{1.4cm}<{\centering} @{} m{1.4cm}<{\centering} @{}}
         \toprule
        & \multicolumn{3}{c}{Sequnce 1} & \multicolumn{3}{c}{Sequence 2} & \multicolumn{3}{c}{Sequence 3} \\ 
        & \multicolumn{3}{c}{train/test views: 207/22} & \multicolumn{3}{c}{train/test views: 223/24} & \multicolumn{3}{c}{train/test views: 337/37}\\
        \cmidrule(lr){2-4} \cmidrule(lr){5-7} \cmidrule(l){8-10}
        
        Model & PSNR $\uparrow$ & SSIM $\uparrow$ & LPIPS $\downarrow$ & PSNR $\uparrow$ & SSIM $\uparrow$ & LPIPS $\downarrow$ &  PSNR $\uparrow$ & SSIM $\uparrow$ & LPIPS $\downarrow$ \\ 
        \midrule
        NeRF\cite{mildenhall2021nerf} & 26.80 & 0.83 & 0.38 & 24.55 & 0.84 & 0.46 & 25.76 & 0.84 & 0.45 \\
        Mip-NeRF 360\cite{barron2022mip} & 27.17 & 0.84 & 0.38 & 24.58 & 0.84 & 0.45 & 26.10 & 0.86 & 0.44 \\
        DVGO\cite{sun2022improved} & 14.28 & 0.70 & 0.64 & 15.10 & 0.74 & 0.64 & 18.76 & 0.80 & 0.55 \\
        Thermal-NeRF w/o pose\cite{wang2021nerf} & 26.78 & 0.88 & 0.30 & 23.36 & 0.85 & 0.39 & 25.59 & 0.87 & 0.36 \\
        Thermal-NeRF & \bf{32.31} & \bf{0.89} & \bf{0.29} & \bf{25.88} & \bf{0.85} & \bf{0.35} & \bf{26.78} & \bf{0.87} & \bf{0.35}\\
        \bottomrule
    \end{tabular}
    \label{tab3}
\end{table*}

\subsection{Self-Collected IR Dataset}
To assess the Thermal-NeRF with IR cameras, due to the current lack of IR datasets for NeRF, we built a new dataset utilizing the Optris PI 450i IR camera and will make it publicly available. The ground truth camera poses were recorded using VICON, a Motion capture (MoCap) system. The camera is equipped with several infra-reflective markers to form a rigid body, positioning it within the MoCap's coordinate system $W$. To increase the precision of camera pose, we conducted a hand-eye calibration, achieving the transformation from $T^{W}_{R}$ to $T^{W}_{C}$, $T^{W}_{R}$ represents the pose of the rigid body, and $T^{W}_{C}$ represents the pose of the camera, both relative to the MoCap system's coordinate system $W$, where $R$ and $C$ correspond to the coordinate systems of the rigid body and the camera, respectively. And for every image $I_k$ and the corresponding pose $T_k$ captured by the MoCap systems, we used spherical linear interpolation for the rotational component of $T_k$, and linear interpolation for the translational component of $T_k$. 

Our IR dataset features three distinct scenarios, each designed to replicate conditions similar to those in fire-related environments: environments with low lighting, fluctuating lighting and smoke, illustrated by Fig. \ref{data_ex}. We simulated thermal sources using heated objects like cups of hot water and heat-emitting electronic devices. Under these varied conditions, our IR cameras demonstrated consistent performance, in contrast to RGB cameras, which experienced diminished functionality in the visible spectrum. For comprehensive scene coverage, we performed 360-degree photography around each indoor scenario, comprising four sequences per scenario, with each sequence containing between 200 to 300 images. Our dataset will be open-sourced for further research.

\subsection{Experimental Setup}

Our experimental framework encompasses novel view synthesis and 3D object reconstruction to comprehensively evaluate the effectiveness and robustness of our proposed method. Additionally, we have verified the critical importance of pose refinement in enhancing the performance of our approach. In assessing novel view synthesis, we employ widely recognized evaluation metrics, including the Peak Signal-to-Noise Ratio (PSNR) \cite{eskicioglu1995image}, SSIM\cite{hore2010image}, and the Learned Perceptual Image Patch Similarity (LPIPS)\cite{zhang2018unreasonable} using VGGNet\cite{simonyan2014very}, to ensure a thorough and objective assessment. Notably, we intentionally transform IR images into pseudo-color representations to improve our analysis by highlighting features that are less visible in grayscale, the jet colormap array is adopted for color conversion, this conversion process does not impact the assessment outcomes and aligns with conventional NeRF evaluation standards. For the 3D object reconstruction aspect, our focus is on reconstructing thermal objects within the scene, we utilize the marching cube method for mesh extraction\cite{lorensen1998marching}, enabling a qualitative evaluation of the reconstructed objects.

\begin{figure*}[thpb]
    \centering
    \begin{minipage}[c]{.01\textwidth}
        \centering
        \rotatebox[origin=c]{90}{\tiny{Low-Light}}
    \end{minipage}%
    \begin{tabular}{c}
        \begin{minipage}{0.96\linewidth}
        \centering
        \includegraphics[width=0.16\linewidth]{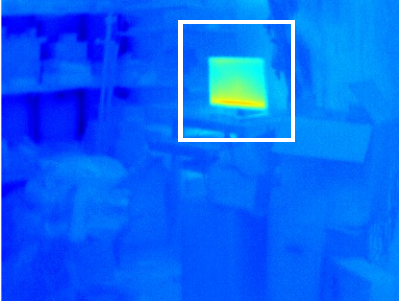}
        \includegraphics[width=0.16\linewidth]{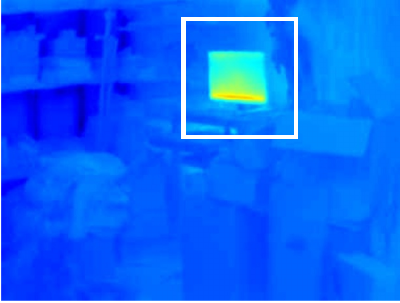}
        \includegraphics[width=0.16\linewidth]{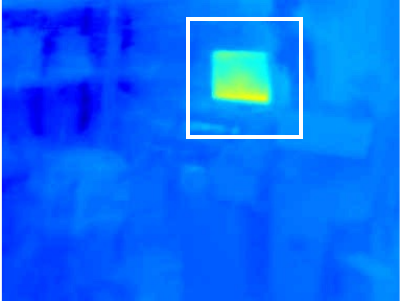}
        \includegraphics[width=0.16\linewidth]{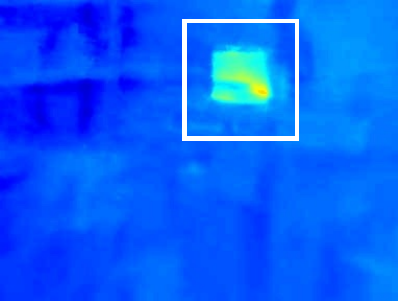}
        \includegraphics[width=0.16\linewidth]{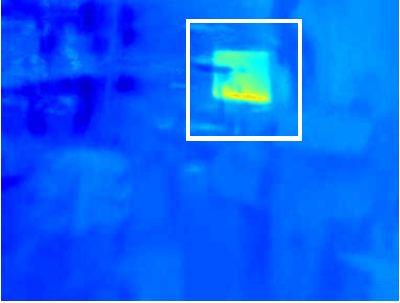}
        \includegraphics[width=0.16\linewidth]{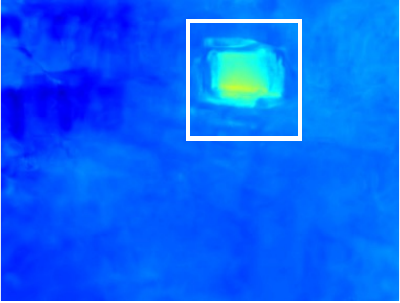}
        \end{minipage} \\
        
    \end{tabular}

    \vspace{1mm}
    \begin{minipage}[c]{.01\textwidth}
        \centering
        \rotatebox[origin=c]{90}{\tiny{Light-Changing}}
    \end{minipage}%
    \begin{tabular}{c}
        \begin{minipage}{0.96\linewidth}
        \centering
        \includegraphics[width=0.16\linewidth]{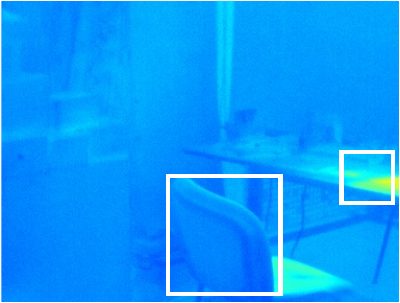}
        \includegraphics[width=0.16\linewidth]{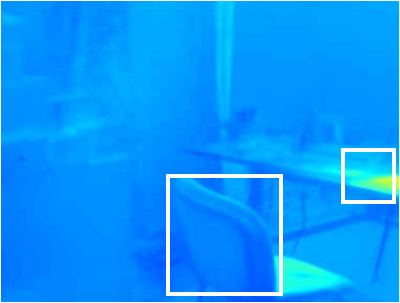}
        \includegraphics[width=0.16\linewidth]{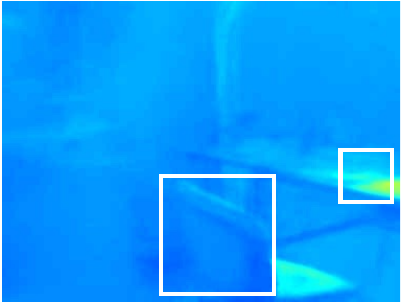}
        \includegraphics[width=0.16\linewidth]{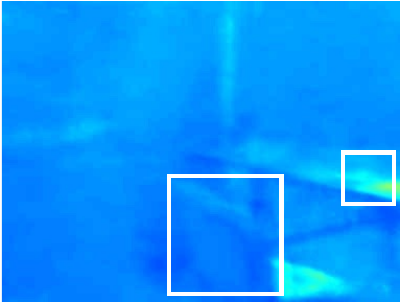}
        \includegraphics[width=0.16\linewidth]{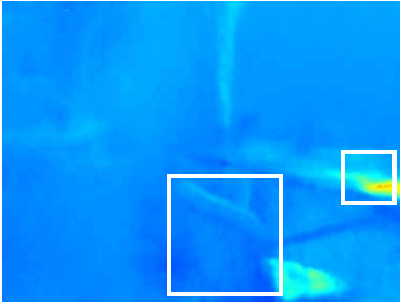}  
        \includegraphics[width=0.16\linewidth]{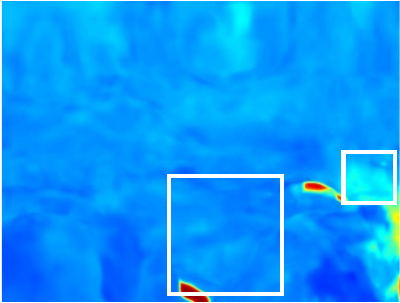}  
        \end{minipage} \\ 
    \end{tabular}

    \vspace{1mm}
    \begin{minipage}[c]{.01\textwidth}
        \centering
        \rotatebox[origin=c]{90}{\tiny{~~~~~~Smoky}}
    \end{minipage}%
    \begin{tabular}{c}
        \begin{minipage}{0.96\linewidth}
        \centering
        \includegraphics[width=0.16\linewidth]{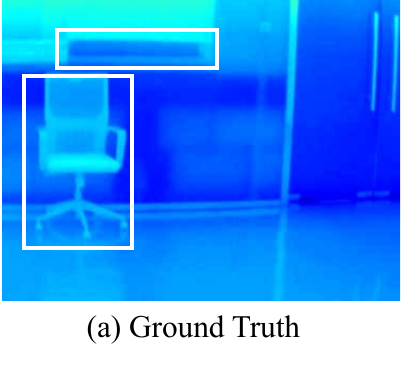} 
        \includegraphics[width=0.16\linewidth]{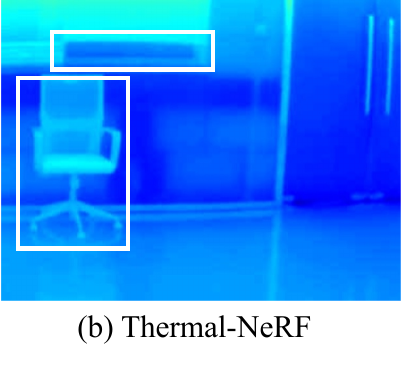} 
        \includegraphics[width=0.16\linewidth]{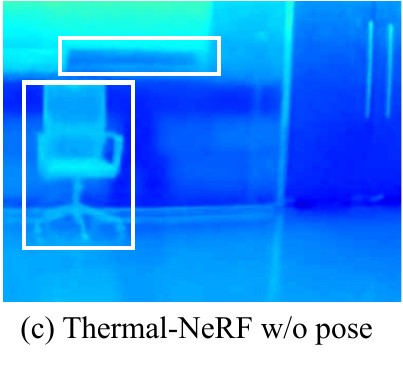} 
        \includegraphics[width=0.16\linewidth]{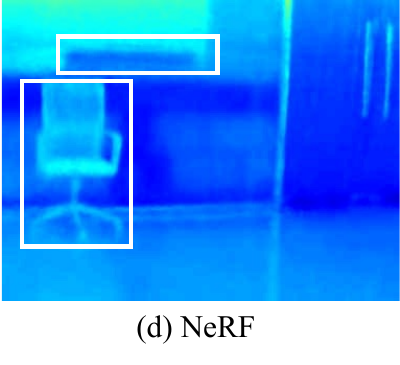} 
        \includegraphics[width=0.16\linewidth]{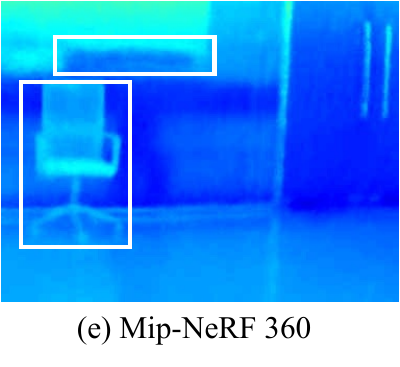} 
        \includegraphics[width=0.16\linewidth]{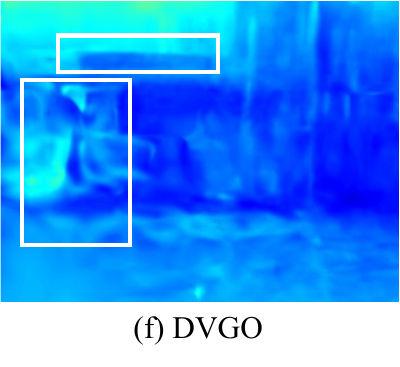} 
        \end{minipage} \\
    \end{tabular}

    \caption{Qualitative evaluation on self-collected sequences under challenging conditions. To enhance the visual assessment of the effects, IR images are intentionally transformed into pseudo-color representations by employing the jet colormap array for the color conversion process. Model: original NeRF\cite{mildenhall2021nerf}, Mip-NeRF 360\cite{barron2022mip}, DVGO\cite{sun2022improved}, and Thermal-NeRF. Specifically, we present results of Thermal NeRF without pose refinement to underscore the significance of pose optimization in achieving optimal outcomes.}
    \label{qualita}
\end{figure*}

\subsection{Novel View Synthesis} 
\label{nvs}
We employ both quantitative and qualitative assessments to evaluate the effectiveness of Thermal-NeRF of novel view synthesis experiments. Quantitative and qualitative analysis, leveraging three challenging scenes from our custom dataset, is succinctly presented in Tables \ref{tab1}, \ref{tab2}, \ref{tab3} and Fig. \ref{qualita} respectively. In our findings, Thermal-NeRF excels at capturing thermal intensities with remarkable precision, crucial for comprehensive thermal analysis. Specifically, our experiments shed light on integrating pose refinement with structural thermal constraint yields superior outcomes, significantly outperforming the results achieved by applying either technique in isolation, which improves image quality and enables Thermal-NeRF to attain its best performance. In contrast to competing models, Thermal-NeRF produces images that are sharp and clear scenes including the details, avoiding the common issue of blurriness. While models like original NeRF and Mip-NeRF 360 capture essential scene features, they struggle with achieving clear images. DVGO, aiming for faster rendering speeds, compromises on rendering accuracy. The superiority is also consistently reflected in our PSNR metrics, where our model surpasses others. Our approach also achieves superior structural and textural accuracy, essential for maintaining scene integrity, as indicated by the SSIM scores. In terms of perceptual alignment with human vision, especially in understanding the subtleties of thermal images, our model shows notable advancements. The LPIPS scores highlight this improvement, demonstrating our model's enhanced ability to interpret semantic details more effectively.

Moreover, although the PSNR values for Thermal NeRF might occasionally fall slightly below those of the more time-consuming methods like NeRF and Mip-NeRF 360 when pose refinement is not used, this phenomenon can be rationalized by considering the distinct focus of our method. PSNR is a metric that emphasizes pixel-level differences, which does not necessarily lead to better visual outcomes. In contrast, our approach prioritizes the visual quality of the images, see Fig. \ref{qualita}, as proven by the superior LPIPS, SSIM scores. These metrics assess structural integrity and perceptual similarity, respectively, and are more aligned with human visual perception by prioritizing aspects that contribute to a visually appealing image over mere pixel accuracy. This underscores the importance of the integration of structural constraints, which are key to enhancing rendering quality.

\subsection{3D Object Reconstruction} 
\begin{figure*}[thpb]
    \centering
    \includegraphics[width=0.95\linewidth]{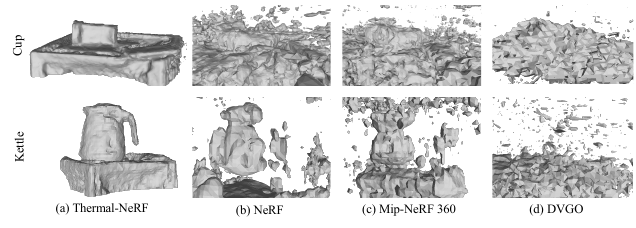}
    \caption{Examples of mesh reconstruction of heat source objects in a scene. Explicit meshes of cup and kettle exported by Thermal-NeRF and other models are shown respectively. }
    \label{mesh}
\end{figure*}

To assess Thermal-NeRF's efficacy in 3D reconstruction, we employ marching cube algorithm for mesh extraction of localized heat sources within the scenes, the specific parameters of the algorithm are adjusted based on each model. In our evaluation, we tested Thermal-NeRF against models including original NeRF, Mip-NeRF 360 and DVGO. The resulting meshes, simulating heat sources with objects like a cup and a kettle containing hot water, are flawlessly shaped and hole-free, underscoring the model's precision, as depicited in (a) of Fig. \ref{mesh}. Notably, while NeRF and Mip-NeRF 360 demonstrated satisfactory performance at the image level as discussed in Sec. \ref{nvs}, their corresponding meshes were similarly plagued by significant noise. This underscores their limitations in precisely capturing the scene's depth information.

We conjecture that the inherent sparsity and low contrast typical of IR images pose significant obstacles for general models, particularly affecting their ability to accurately estimate density. This often results in outputs marred by considerable noise. In contrast, by incorporating structural constraints, Thermal-NeRF adeptly navigates these challenges. These constraints enable the model to concentrate on areas with pronounced thermal activity, effectively prioritizing the reconstruction of critical heat-emitting objects within a scene. 

\begin{table}[ht]
    \centering
    \caption{Quantitative Ablation Study On the Self-collected Dataset.}
    \label{table4}
    \begin{tabular}{lccc}
    \toprule
        Thermal-NeRF method & PSNR $\uparrow$ & SSIM $\uparrow$ & LPIPS $\downarrow$ \\
    \midrule
    w/o thermal mapping & 22.64 & 0.71 & 0.69 \\
    w/o structural thermal constraint & 24.70 & 0.79 & 0.64 \\
    Thermal-NeRF (default) & \bf{30.52} & \bf{0.88} & \bf{0.41}\\
    \bottomrule
    \end{tabular}
    \label{tab4}
\end{table}

\subsection{Ablation Studies}
\label{abla}
\begin{figure}[t]
    \centering
    \includegraphics[width=0.99\columnwidth]{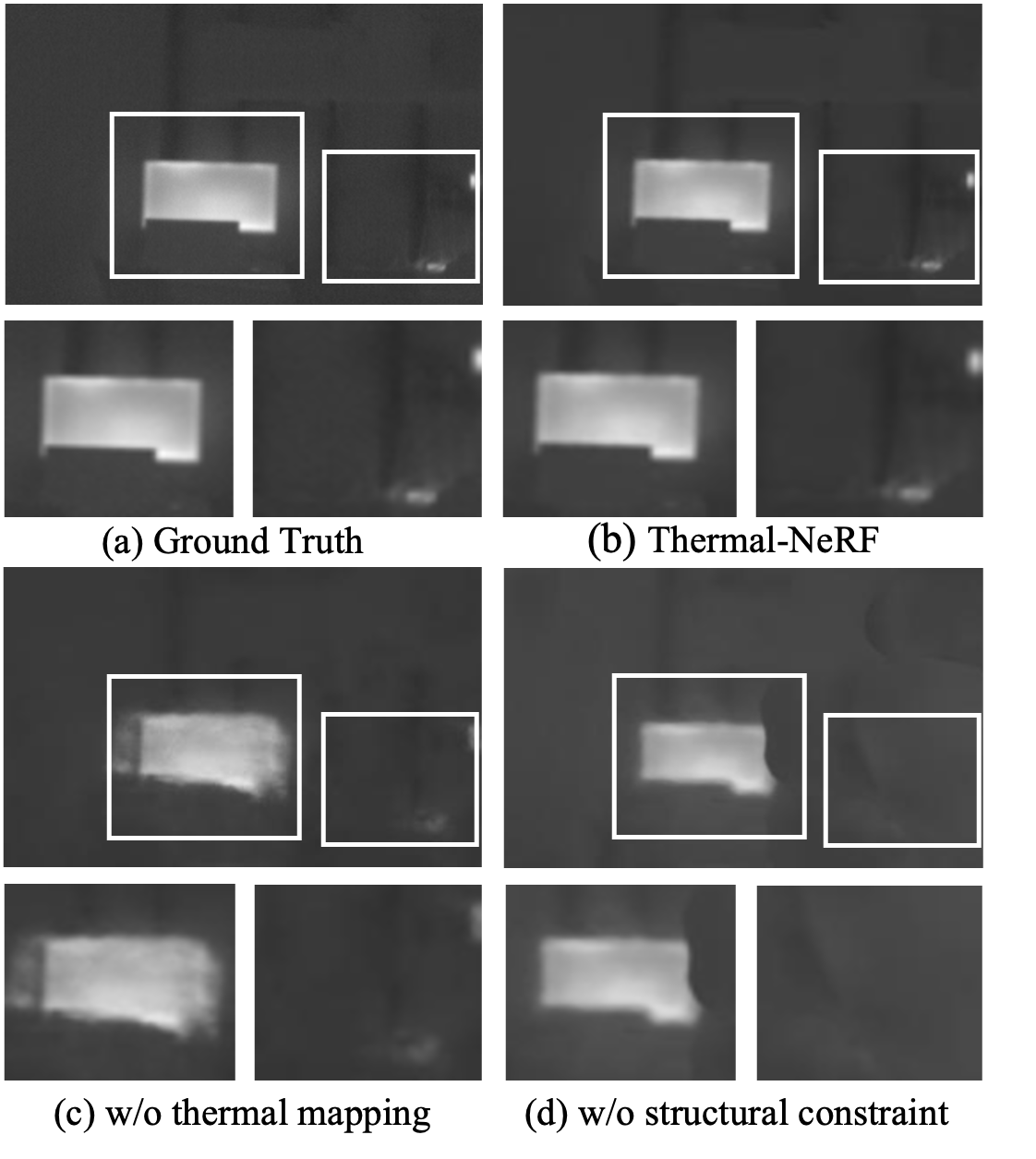}
    \caption{Ablation qualitative example. Here we show renderings from different Thermal-NeRF ablation variants. The full model produces the best results. We zoom in on crops to highlight differences in the rendered images.}
    \label{ablation}
\end{figure}

In our study, we conduct a series of ablation experiments by individually altering components of our method. We evaluate the performance impact of these changes using average image metrics across six self-collected sequences in three challenging environments. The results of these experiments are illustrated in the accompanying Tab. \ref{tab4} and Fig. \ref{ablation}.

Our first ablation involves the mapping method. We replace our thermal mapping approach with a trivial pixel mapping method, which involves scaling the pixel values of single images from a 16-bit format to an 8-bit format using min-max normalization. This modification results in a significant drop in performance, characterized by blurring and ghosting effects. This outcome is anticipated, as such a change violates the thermal consistency inherent in IR imaging. Without mapping the images to a uniform temperature range, pixel values for the same spatial point vary with the viewpoint, leading to inconsistencies. This effect can be shown in Fig. \ref{ablation}, where the trivial pixel mapping results in a substantially different pixel distribution compared to results applying thermal mapping.

Subsequently, we ablate the thermal structural constraint component of our method, Equation \ref{eq10}. This alteration results in a degradation of image quality, with noticeable ghosting in detail features and overall blurring. It becomes clear that pixel-level loss is insufficient for IR imaging, which typically have low contrast. The introduction of structure thermal constraint significantly mitigates artifacts in uniformly colored areas and effectively recovers detailed features in areas with concentrated heat, see Fig. \ref{ablation}.

\section{CONCLUSIONS}
In this study, we introduce Thermal-NeRF, the first approach for reconstructing neural radiance fields exclusively from IR imaging, particularly beneficial in visually degraded robotics scenarios, including low-light, light-changing, and smoky environments. Thanks to the proposed combination of thermal mapping and a structural thermal constraint, Thermal-NeRF outperforms existing methods on our custom IR dataset, delivering improved quality in both image rendering and mesh reconstruction of heat sources. Thus, this paper extends the spectrum of practical IR-based techniques with a 3D representation learning approach. Future work could involve integrating depth supervision to enable comprehensive scene-level reconstruction. 

\bibliographystyle{IEEEtran}
\bibliography{BibFile}

\end{document}